\documentclass[10pt,twocolumn,letterpaper]{article}

\usepackage{iccv}
\usepackage{times}
\usepackage{epsfig}
\usepackage{graphicx}
\usepackage{amsmath}
\usepackage{amssymb}
\usepackage{algorithm}
\usepackage{algorithmicx}
\usepackage{subfigure}

\usepackage[pagebackref=true,breaklinks=true,letterpaper=true,colorlinks,bookmarks=false]{hyperref}

\iccvfinalcopy 

\def\iccvPaperID{1342} 

\ificcvfinal\fi
\begin{document}

\title{Planecell: Representing the 3D Space with Planes}

\author{Lei Fan, Ziyu Pan\\
Sun Yat-sen University\\
{\tt\small \{fanl7,panzy6\}@mail2.sysu.edu.cn}
\and
Long Chen, Kai Huang\\
Sun Yat-sen University\\
{\tt\small \{chenl46,huangk36\}@mail.sysu.edu.cn}
}

\maketitle

\begin{figure*}[t]
	\centering
	\includegraphics[width=1.0\linewidth]{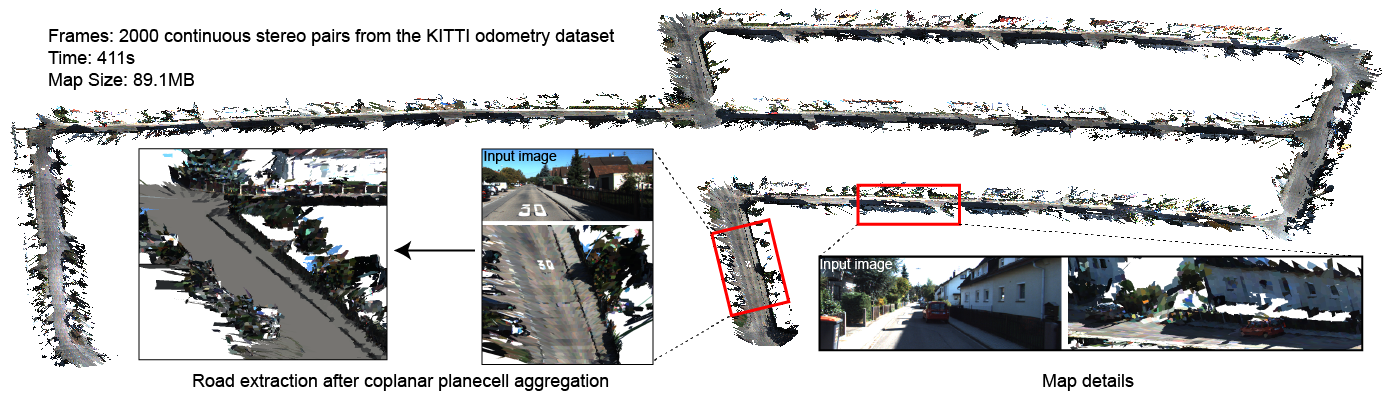}
	\caption{The map reconstructed by proposed method. The result shows three advantages: (a) the ability of dealing with large-scale reconstruction; (b) the low-loss of accuracy (detailed quantitative evaluation is processed on KITTI stereo dataset compared to ground truth); (c) the low requirements of both time and memory.}
	\label{fisrtView}
\end{figure*}
\begin{abstract}
	Reconstruction based on the stereo camera has received considerable attention recently, but two particular challenges still remain. The first concerns the need to aggregate similar pixels in an effective approach, and the second is to maintain as much of the available information as possible while ensuring sufficient accuracy. To overcome these issues, we propose a new 3D representation method, namely, planecell, that extracts planarity from the depth-assisted image segmentation and then projects these depth planes into the 3D world. An energy function formulated from Conditional Random Field that generalizes the planar relationships is maximized to merge coplanar segments. We evaluate our method with a variety of reconstruction baselines on both KITTI and Middlebury datasets, and the results indicate the superiorities compared to other 3D space representation methods in accuracy, memory requirements and further applications. 
\end{abstract}

\section{Introduction}
3D reconstruction has been an active research area in the computer vision community, which can be used in numerous tasks, such as perception and navigation of intelligent robotics, high precision mapping, and online modeling. Among various sensors that can be used for 3D reconstruction, stereos cameras are popular for offering advantages in terms of being low-cost and supplying color information. Many researchers have improved the precision and speed of self-positioning and depth calculation algorithms to enable better reconstruction, but few have attempted to change the basic map representation method which determines the upper bound of reconstructions. Current approaches including point-based or voxel-based representations are confronted with problems, such as significant redundancy, ambiguities, and memory requirements. To overcome these limitations, we propose a new representation method named \emph{planecell}, which models planes to deliver geometric information in the 3D space.


It is a classical approach to representing the 3D space with a preliminary point-level map. The point-based representations usually suffer a tradeoff of density and efficiency. Many approaches~\cite{osman2016patches,agrawal2001probabilistic,hane2013joint} have been developed to address this issue, \ie, to merge similar points in the 3D reconstruction results for both indoor and outdoor scenes. The current leading representation method, called the voxel map~\cite{bhotika2002probabilistic,whelan2013robust,osman2016patches,vineet2015incremental}, is designed to give each voxel grid an occupancy probability, and then aggregates all points within a fixed range. However, dense reconstructions using regular voxel grids are limited to reach small volumes because of their memory requirements.

Previous studies have adopted the plane prior both in stereo matching~\cite{yamaguchi2014efficient} and reconstruction~\cite{osman2016patches}. Ulusoy \etal. presented a Markov random field model in the former work~\cite{osman2016patches} for volumetric multi-view 3D reconstruction. The model uses large 3D surface patches that can be encoded as probabilistic priors. Deriving primitives in the model raises the complexity and restricts further applications. Methods that derive the planarity parameter or use the plane model are based on the fact that the world we live in is mostly composed of plane structures, especially in man-made environments.

In this paper, we propose a novel representation method that differs from existing approaches by mapping the 3D space with basic plane units, which is called planecell for it resembles cells to a living being. The proposed method utilizes a general function to represent a group of points with similar geometric information, i.e., belong to the same plane by a depth-aware superpixel segmentation, and these planes are projected into the real-world coordinates after plane-fitting with depth values. The standardized representation promotes memory efficiency and provides convenience for following computations, such as surface segmentation and distance calculation. Our method starts from extracting planecells from 2D images by superpixelizing the input image following the hierarchical strategy of SEEDS~\cite{van2012seeds} and converts them into a 3D map. The planecells are then merged by modeling a Conditional Random Field (CRF) formulation. Unlike existing surface estimation methods, the aggregation of coplanar units applying proposed CRF formulation only needs to refer to the properties of each planecell, which dramatically reduces required computations. The proposed representation is motivated by the planar nature of the environment. The input to our method is a color reference image and the corresponding depth map, and the output is a plane-based 3D map.
In our experiments, we evaluate different input disparity images from various matching algorithms in our experiment to demonstrate the adaptiveness of our technique, and we compare our planecell with existing popular 3D space representation approaches. 

The detailed contributions of this paper are as follows: (a) We propose a novel plane-based 3D map representation method that demonstrates remarkable accuracy and has enhanced the space perception abilities. (b) A CRF model that aggregates coplanar planecells in 3D space is proposed. (c) The accuracy and efficiency of our representation method are studied by comparison to existing popular approaches. We also show the accessibility of applications include but not limited to road extraction and obstacle avoidance on our planecell representation in the experiment. In practice, this objective can be optimized on a single core in as little as 0.2 seconds for about 700 planecells. To further aggregate coplanar planes requires only $0.1$s per frame. More detailed results can be found at \url{http://www.carlib.net/planecell.html}.

\begin{figure*}[t]
	\centering
	\includegraphics[width=1.0\linewidth]{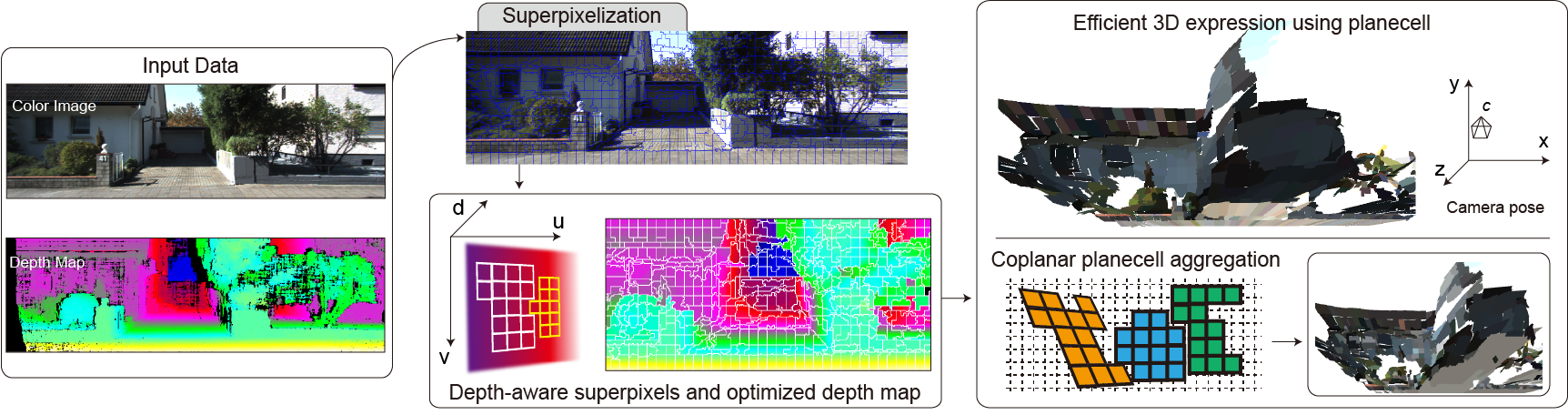}
	\caption{Overview of our 3D representation method: planecell.}
	\label{overView}
\end{figure*}

\section{Related Work}
Basic 3D map representation methods using an image pair are inheritors of various stereo matching algorithms~\cite{yamaguchi2014efficient,hirschmuller2007stereo,zbontar2016stereo,einecke2010two,luo2016efficient,guney2015displets}. Point-based 3D reconstruction methods directly transforming stereo matching results lack structural representations. Recent point-level online scanning~\cite{zollhofer2014real} produces a high-quality 3D model of small objects with the geometric surface prior, which is simpler to operate than strong shape assumptions. For large-scale reconstructions, sparse point-based representations are mainly used for their quality and speed. The point-based maps embedded in the system~\cite{einecke2010two} is designed for real-time applications, such as localization. Different features have been developed for this purpose. For example, the ORB feature matching~\cite{rublee2011orb} is designed for fast tracking via a binary descriptor. Adopting denser point clouds in the mapping is challenging because it involves managing millions of discrete values. 


The heightmap is a representation adopting 2.5D continuous surface representations, which shows its advancement modeling large buildings and floors. Gallup \etal proposed an $n$-layer heightmap~\cite{gallup20103d} to support more complex 3D reconstruction of urban scenes. The proposed heightmap enforced vertical surfaces and avoided major limitations when reconstructing overhanging structures. The basic unit of heightmap is the probability occupancy grid computed by the bayesian inference, which could compress surface data efficiently but is also lossy of point-level precision. 

Recent studies on voxelized 3D reconstruction focus on infusing primitives into the reconstructions~\cite{liu2015higher,osman2016patches,fraundorfer2006piecewise,chauve2010robust} or utilizing scalable data structures to meet CPU requirements~\cite{vineet2015incremental}. Dame \etal proposed a formulation which combines shape priors-based tracking and reconstruction. The map was represented as voxels with two parameters including the distance to the closest surface and the confidence value. Nonetheless, the accuracy of volumetric reconstruction is always limited to itself, and re-estimating object surfaces from voxels or 3D grids lead to ambiguities. 

Planar nature assumptions have been applied to both the reconstruction~\cite{liu2015higher} and depth recovery from a stereo pair~\cite{yamaguchi2014efficient,guney2015displets,vogel20153d}. For surface reconstruction or segmentation, Liu \etal~\cite{liu2015higher} partitioned the large-scale environment into structural surfaces including planes, cylinders, and spheres using a higher-order CRF. A bottom-up progressive approach is adopted alternately on the input mesh with high geometrical and topological noises. Adopting this assumption, we present a new representation method of 3D space, which is composed of planes with pixel-level accuracy.

\section{System Overview}
As shown in Fig.~\ref{overView}, the input to the system is a combination of a color image and the disparity map. We use a depth-aware superpixel segmentation method with an additional \emph{depth term}. A hill-climbing~\cite{van2012seeds} superpixel segmentation method is applied to the color image with a \emph{regularization term} to reduce the complexity. The disparity map is pre-calculated with stereo matching algorithms. Sparse results produced by fast algorithms can still be the input, as we utilize random sampling to omit the effect of outliers during plane-fitting. The boundaries of the segmentation are further updated after plane functions have been assigned to each segment. The superpixels are the basic elements of the mapping process. We extract the vertexes of each plane and then convert them into the camera coordinate system. For existing 3D planes, we aggregate those whose spatial relationship are planarity while minimizing the total energy function.

As the core of our algorithm is independent of the choice of 3D knowledge acquisition, we can alternate the input into the ground truth from laser scanners. However, stereo camera has the advantages of being a low-cost solution for obtaining both depth and color information. In the next section, we primarily describe the process of using stereo pairs as inputs, and we impose the SGM to obtain the depth map.

\section{Representing the 3D Space with Planecells}
The planecell is the basic unit representing geometric information of objects in the 3D space. Each planecell is a combination of pixels from the color image and uses a joint plane function to deliver their positions. The shape of each planecell is a polygon, which enables us to define their boundaries by vertexes. The planecells are adopted by two main processes of stereo matching (introduced with SGM method) and superpixel segmentation, which will be explained separately.

\subsection{Depth Map Calculation with SGM}
The proposed method first calculates a semi-dense disparity map on the input image $\mathcal{I}_L$ with a kind of SGM method that combines both the Census transformation and gradient information. Denote $\mathcal{T}(.)$ as the descriptor of the Census transformation and $\mathcal{H(.,.)}$ as the Hamming distance between two descriptors. Let $\mathcal{G}(.,.)$ be the directional gradient in the image. The matching cost between the pixel $\rm{p}$ in the left image and pixel ${\rm{q}}=(\rm{p}_x-d_{\rm{p}}, \rm{p}_{y})$ on the epipolar line $e(\rm{q})$ in the right image is defined as
\begin{equation}
\begin{split}
C(\rm{p},\rm{q}) =& \mathcal{H}(\mathcal{T}_{L}(\rm{p}), \mathcal{T}_{R}(\rm{q})) \\
&+\lambda_{grad}|\mathcal{G}_L(\rm{p},e(\rm{q}))-\mathcal{G}_R(\rm{q},e(\rm{q}))|
\end{split}
\end{equation}
Applying the minimum cost path $L_r$ aggregated in direction $r$ with penalties for discontinuities, the final disparity $d$ of pixel $\rm{p}$ is calculated as
\begin{equation}
d = \rm{argmin}\{\sum_{r}L_r(p,d)\}
\end{equation}

\subsection{Planecells Extraction}
We utilize superpixel segmentation methods to adopt basic planecells from the color reference image. For a color image $\mathcal{I}$, the superpixel segmentation $ S=\{S_1,...,S_k\}$ has the following properties
\begin{equation}
S_i \cap S_j=\emptyset \quad and \quad \mathcal{I}=\bigcup_{Si\in S} \quad
\end{equation}
Throughout the literature, various superpixel algorithms are graph-based methods that aggregate similar pixels are belonging to the same object. This property helps to distinguish planes in the region of interest initially, and superpixel segmentation leaves no holes in the input reference image, which also benefits the 3D map representation. However, the boundary of each superpixel is always unnecessarily heavy and complicated, especially for urban scenarios with structural objects, which increases the computational and storage demands. To address this issue, we propose an improved superpixel method based on the prior work~\cite{van2012seeds}. We define the smallest unit of boundary update as a block $\rm{b}$ instead of a pixel. In practice, the size of $\rm{b}$ is related to the levels $l$ of the hill-climbing algorithm and the number of superpixels $|S|$, which is set as $\lfloor\sqrt{\frac{\mathcal{I}}{|S|}}\rfloor/l$ pixels.

\begin{figure}[t]
	\centering
	\subfigure[Regularization term $R$]{
		\includegraphics[width=0.35\linewidth]{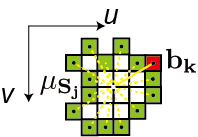}
	}
	\subfigure[Depth term $D$]{
		\includegraphics[width=0.4\linewidth]{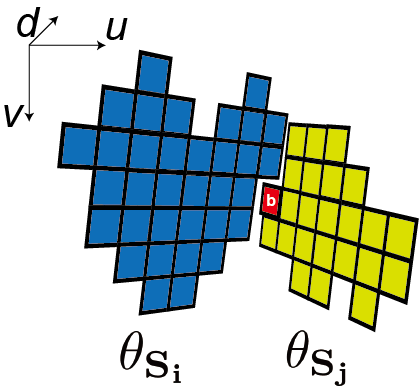}
	}
	\caption{Beyond color distribution term, we add two terms according to the location and depth of blocks which are the minimum changing units during boundary update.}
	\label{superpixelization}
\end{figure}
\subsection{Superpixel Energy Function}
The superpixel segmentation is bounded by the maximization of the energy function, which is defined as the sum of three terms. The energy comprises a color term $H(S)$ based on the histogram of the color space, a regularization term $R(S)$, and a depth term $D(S)$: 
\begin{equation}
\label{SuperEnergy}
E(S)=H(S)+\lambda_{reg}R(S)+\lambda_{depth}D(S)
\end{equation}
where $\lambda_{reg}$ and $\lambda_{depth}$ are two balancing parameters.\\
\textbf{Color term:} The color term $H(s)$ measures the color distribution of the superpixels and inclines toward superpixels with color histograms that drop into similar bins. With the image segmentation $S$, the color term is formulated as
\begin{equation}
H(S)=\sum_{S_i\in S}\sum_q^Q h_{S_i}(q)^2
\end{equation}
where $q$ denotes the histogram bin and $h(.)$ is the number of pixels in the bin. It is not difficult to infer that $H(S)$ reaches its maximum if and only if each histogram is placed in the same bin. Nonetheless, the quality of this evaluation of color is related to the bin size, i.e., the sensibility of color declines when the number of neighboring colors in a single bin is large.\\
\textbf{Regularization term: } The regularization term $R(S)$~(see Fig.~\ref{superpixelization}(a)) constrains the superpixels to be standard, encouraging straight boundaries. Let $\mu_{S_i}$ be the center of segment $i$, $\mathcal{B}_{i,j}$ be the set of boundary blocks between segment $i$ and segment $j$, and $\mathcal{N}_{S_i}$ be the set of adjacent segments of segment $i$. The regularization term is given by
\begin{equation}
\label{regular}
\begin{split}
R(S)=&\sum_{S_i\in S}\sum_{S_j\in \mathcal{N}_{S_i}}\sum_{b_k\in \mathcal{B}_{i,j}}\|\mu_{S_i}-\mu_{S_j}\|_2^2\\
&-\|\mu_{S_i}-b_k\|_2^2-\|\mu_{S_j}-b_k\|_2^2
\end{split}
\end{equation}
The value of $R(S)$ is maximized when all blocks on the boundary have the same distances to the neighboring superpixels.\\
\textbf{Depth term:} The depth term~(see Fig.~\ref{superpixelization}(b)) comes into effect after the plane function of each segment has been obtained. We denote the plane function of $S_i$ as $\theta_{S_i}$ which equals $(A_i, B_i, C_i)$. The depth distance between a block and neighboring segments is estimated by measuring the difference in the average block depth and the estimated depth generated by the plane function. The formula for $D(S)$ is quite similar to $R(S)$ with the two-dimensional coordinate alternatives to the disparity. By applying this term, the segmentation outperforms the former method when the color loses its effect.

\subsection{Plane Function Estimation}
After importing the disparity image, we assign each pixel a label to distinguish whether it is an outlier, i.e., the unmatched pixels. To further identify mismatched pixels, we estimate the plane function by random sampling. The plane-fitting terminates when the number of inliers reaches a target percentage. The difference between the estimated disparity of pixel $\rm{p}$ with segment $i$ and the input disparity is measured as $\theta_{S_i}(\rm{p})-d(\rm{p})$. If this term exceeds a given threshold, the pixel is considered to be an outlier. If no appropriate function can be obtained after a designated number of iterations of $Si$, we omit this segment and re-estimate it after the boundary is updated with depth.

\subsection{Block-level Update}
The proposed method is implemented using a hill-climbing algorithm, which reduces the computational complexity as it allows for faster convergence by changing the size of the initial blocks. Nonetheless, when the updating blocks become bigger, the accuracy decreases. The block size shrinks after the movement of bigger blocks has finished. At level $l$ of the hill-climbing process, the algorithm proposes a new partitioning $S_l$ with blocks changing to its neighboring superpixel horizontally or vertically. The partitioning process is evaluated by the superpixel energy function (Eq.\ref{SuperEnergy}). In our implementation, to adopt more efficient segmentation, the boundary block alters its label at level $l$ depending on the costs defined below:
\begin{equation}
\begin{split}
&c_{1}(S_i,b_j^{l})=\sum_{q}\min\{h_{S_i}(q),h_{b_j^{l}}(q)\}\\
&c_{2}(S_i,b_j^{l})=\lambda_{reg}\|\mu_{S_i}-b_j^l\|_2^2+\lambda_{depth}(\theta_{S_i}(b_j^{l})-\bar{d_{b_j^l}})^2
\end{split}
\end{equation}
where $\bar{d_{b_j^l}}=\frac{\sum_{p\in b_j^l}d_p}{|b_j^l|}$. Note that $c_{1}$ and $c_{2}$ increase and decrease separately when measuring the same block $b$. $c_{1}$ and $c_{2}$ are evaluated during the boundary blocks updating. As the minimum updating unit of our algorithm is at the block level, we iterate changing the boundaries with the smallest blocks until a valid image partitioning is obtained or the maximum run-time is reached.


\begin{figure}[t]
	\centering
	\subfigure[The vertexes extraction rugulation on the segmentation. Each circle denotes a pixel, and the number is the label of superpixel it belongs to.]{
		\includegraphics[width=0.46\linewidth]{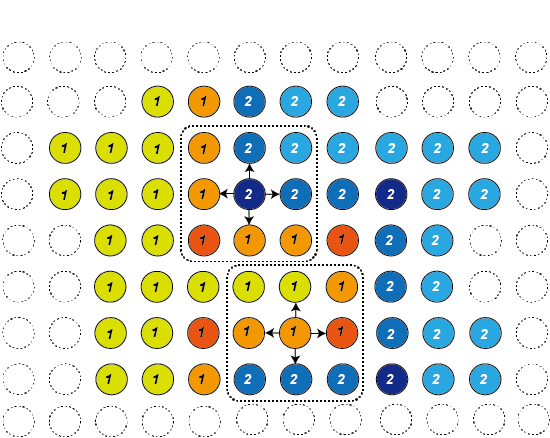}
	}
	\hspace{0.01cm}
	\subfigure[An example with extraction of vertexes. The vertexes of each planecell are marked with the same color.]{
		\includegraphics[width=0.46\linewidth]{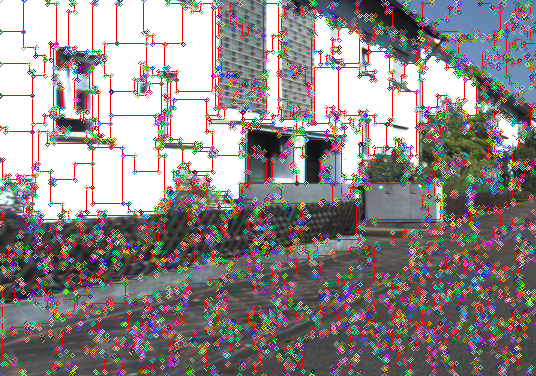}
	}
	\caption{The vertexes extraction.}
	\label{Vertex}
\end{figure}

\section{3D Map Expression}
After producing the partition results with a plane function assigned to each superpixel, we extract the vertex of each segment. The vertex $v_i$ of each segment $i$ is the set of intersections of horizontal and vertical edges. Vertexes require much less memory and computation time than storing all of the plane pixels or edges during 2D-3D conversion. We propose a method of selecting the vertexes on the segmentation results by referring to the count of adjacent pixels that belong to the same superpixel. As demonstrated in Fig.~\ref{Vertex}, the pixel is a $boundary\,pixel$ when three neighboring pixels belong to a different superpixel and two adjacent pixels on the horizontal or vertical line belongs to the same superpixel, and a $vertex\,pixel$ when only one or larger than $3$ neighboring pixels belong to the same superpixel. The result after extracting the vertexes is shown in Fig.~\ref{Vertex}(b).

\subsection{2D-3D Conversion}
The conversion is based on the vertexes. Each vertex set $v_i$ contains $N$ vertexes, and $v_{i,n}$ is composed of variables describing their location in the 2D image and the disparity value estimated with the plane function. Then, for $v_i$ in segment $S_i$, the position in the 3D coordinate system of the left camera can be calculated using the camera's intrinsic parameters and the relative rotation and translation matrix between the stereo camera. We denote the plane function as $\theta '$ after converting. It should be noted that the 2D-3D converting does not cause loss of precision to each pixel.

\subsection{Coplanar Planecells Aggregation with CRF}
The process of aggregating coplanar planecells starts from a plane-based model reconstructed from 2D-3D conversion of vertexes. The target is to assign each planecell with a common label if they fit into a similar geometric primitive in the 3D world. This aggregation reveals higher-level comprehension of the environment, which can be further used in the road extraction and understanding of structures. Prior methods utilizing CRF to merge pixels in the 3D world for the purpose of surface segmentation do not integrate existing knowledge of the color image sufficiently, which requires significant computational resources, especially when dealing with large-scale maps.

\begin{figure}[t]
	\centering
	\includegraphics[width=1.0\linewidth]{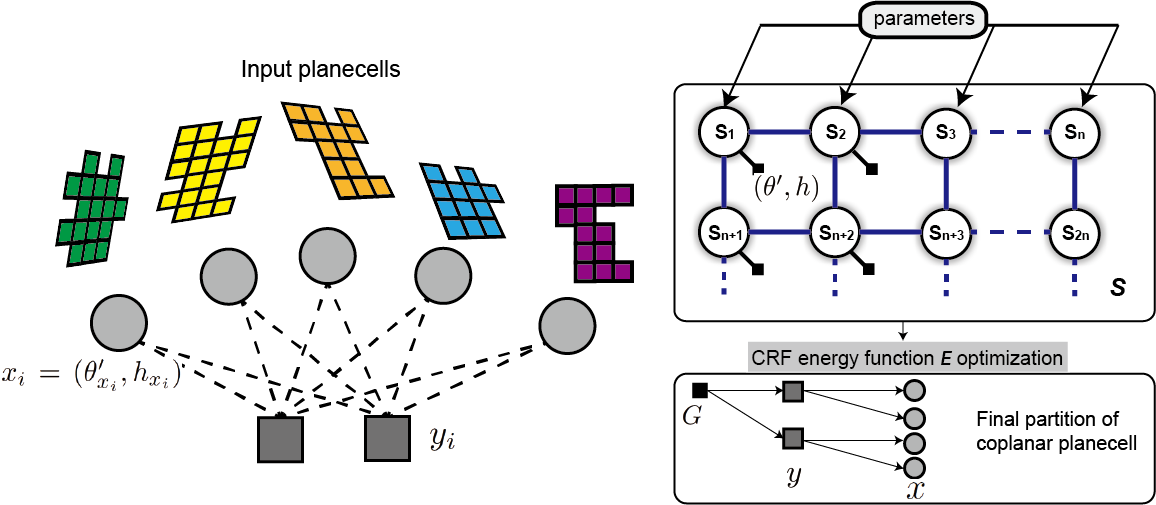}
	\caption{The CRF model.}
	\label{CRF}
\end{figure}
The plane-based map is demonstrated by a set of discrete plane units $\{x_0, x_1, \dots, x_n\}$. The process is then presented as a labeling problem from the CRF model $G$, i.e., assign each unit $x_i$ a label $y_i$ whose value indicates the most probable surface to which it belongs. We denote a tuple $(\theta ', h)$ to describe a plane, where $\theta ’$ is the plane parameters in the camera coordinates and $h$ is the color distribution descriptor. The CRF model is shown in Fig.~\ref{CRF}. The implementation of our process merges coplanar units into a larger surface iteratively until each surface is denoted with a unique tuple. The CRF model at the $t$-th iteration is defined as
\begin{equation}
G^{(t)}=(\bf{x}^{(t)},\bf{y}^{(t)},\bf{c}^{(t)},\bf{o}^{(t)})
\end{equation}
where $\bf{x}^{t}={x_i^{(t)}}$ is the set of nodes denoting the surface units, $\bf{y}$ $={y_i^{(t)}}$ is the labels of $\bf{x}^{(t)}$, $\bf{c}^{(t)}$ is the set of boundaries between each adjacent units, and $\bf{o}^{(t)}$ is a descriptive label of the boundary $\bf{c}^{(t)}$.

The CRF energy function is then formulated as the following (the superscript $^{(t)}$ has been omitted)
\begin{equation}
\label{CRFFunction}
E=\sum_{x_i\in S}\delta_i(y_i) + \sum_{x_j\in \mathcal{N}_{x_i}}\phi_{i,j}(y_i,y_j)+ \sum_{c_i\in \mathcal{B}_{x_i}}\psi_{i}(\bf{y}_c)
\end{equation}
where the potential $\delta_i$ evaluates the color distribution of $x_i$ from the reference image using the histogram of the color space, the pairwise potential $\phi_{i,j}$ measures the difference of depth in the 3D space, which encourages neighboring planecells to belong to the same surface if they are close in both geometric position and pose, and the term $\psi_{i,j}$ technically encodes the boundaries of $x_i$. These potentials are further explained in the following.

The term $\delta_i$ is a unary potential that measures the similarity of the unit and the surface with respect to the color histogram:
\begin{equation}
\sum_{x_i\in S}\delta_i(y_i)=\sum_{x_i\in S}\sum_{q}^{Q}h_{y_i}(h(x_i))
\end{equation}
where $h(x_i) \in \{1,\dots, Q\}$ is the histogram bin of unit $x_i$. This potential increases when the similarity in color rises. The potential $\phi_{i,j}(y_i,y_j)$ is designed to constrain the geometric information, which refers of the $\theta '_i$ in each planecell:
\begin{equation}
\sum_{x_j\in \mathcal{N}_{x_i}}\phi_{i,j}(y_i,y_j)=\sum_{x_j\in \mathcal{N}_{x_i}}\frac{|x_i\cup x_j|}{\theta '_i(x_j)^2+\theta '_j(x_i)^2+1}
\end{equation}
where $\theta '(.)$ denotes the difference value with coordinates of the plane function $\theta '$. Let $\theta_i '= (A_i ',B_i ',C_i ')$ be the plane pose of unit $x_i$. The 3D point $p$ in unit $x_i$ (planecell) obeys $\theta_i ' \cdot p = 0$. The potential $\phi_{i,j}$ reaches its maximum when two units agree in their poses. For the potential $\psi_i$, the formula can be written as
\begin{equation}
\sum_{c_i\in \mathcal{B}_{x_i}}\psi_{i}({\bf{y}_c})=\sum_{c_i\in \mathcal{B}_{x_i}}\frac{|c_i|}{\theta_i '(c_i)^2+1}
\end{equation}

The maximization of Eq.~\ref{CRFFunction} is an NP-hard problem solving a CRF model with various variables. We implement the labeling process using a circular greedy algorithm, which merges units within a given range of variation to the greatest extent. Note that the variation range determines the possibility of planarity between adjacent plane units.

\begin{figure*}[t]
	\centering
	\subfigure[KITTI stereo dataset with non-occlusion grount truth]{
		\includegraphics[width=0.32\linewidth]{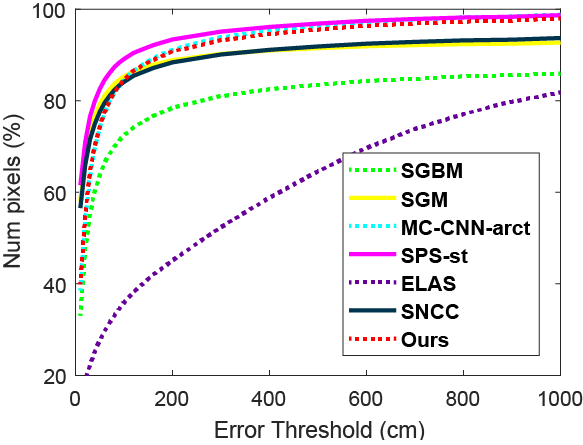}
	}
	\subfigure[KITTI stereo dataset with occlusion-included grount truth]{
		\includegraphics[width=0.32\linewidth]{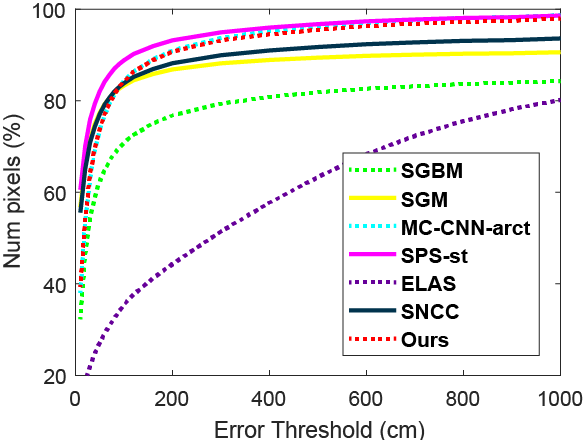}
	}
	\subfigure[Middlebury stereo dataset with occlusion-included grount truth]{
		\includegraphics[width=0.32\linewidth]{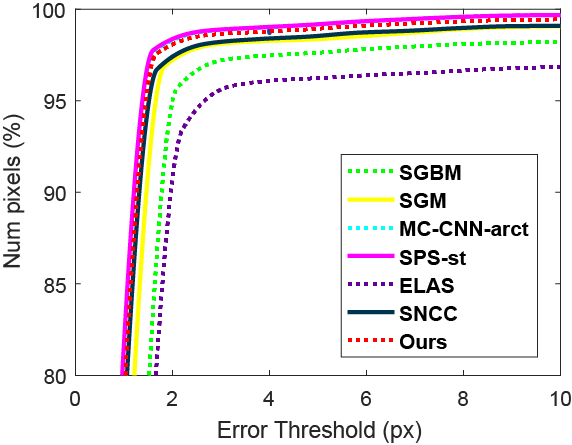}
	}
	\caption{Reconstruction accuracy plots for KITTI and Middlebury stereo datasets.}
	\label{Compare2Point}
\end{figure*}

\begin{figure}[t]
	\centering
	\subfigure[KITTI stereo dataset]{
		\includegraphics[width=0.47\linewidth]{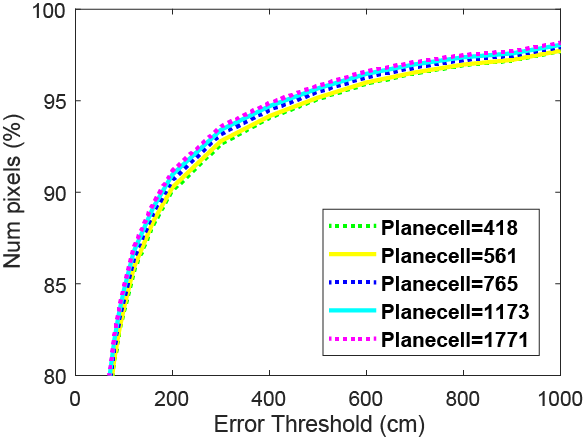}
	}
	\subfigure[Middelbury stereo dataset]{
		\includegraphics[width=0.47\linewidth]{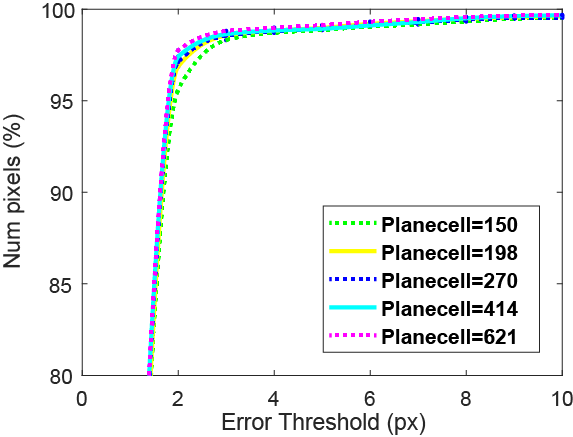}
	}
	\caption{Reconstruction accuracy plots with different number of planecells.}
	\label{Compare2Self}
\end{figure}

\begin{figure*}[t]
	\centering
	\includegraphics[width=1.0\linewidth]{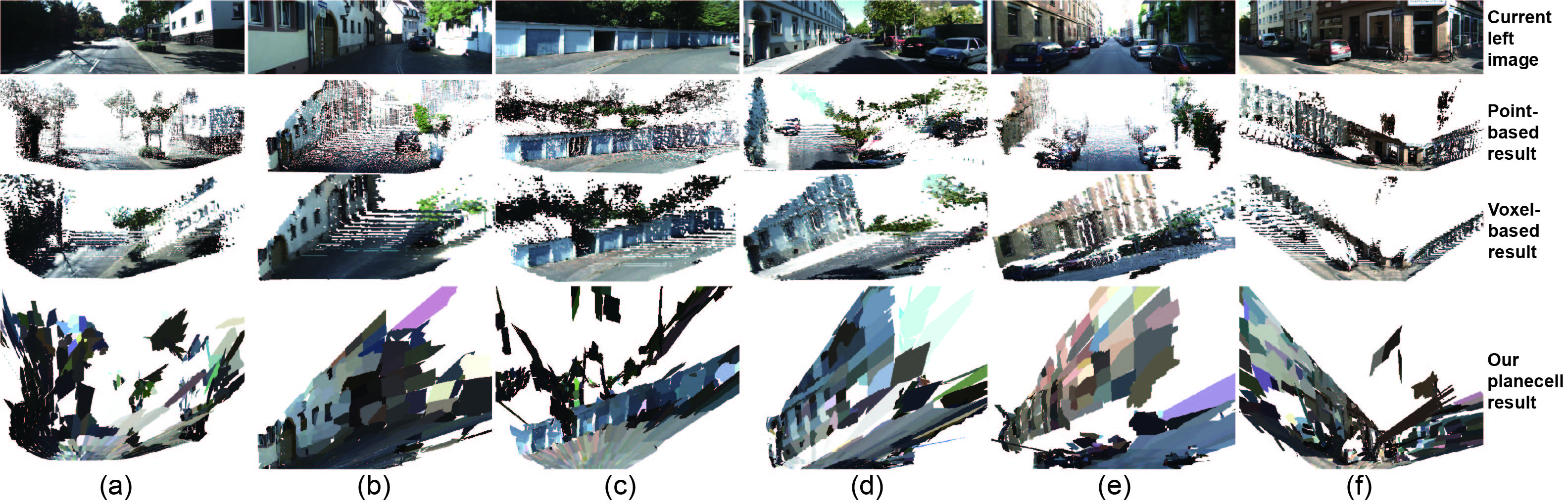}
	\caption{The 3D space representation with point-based, voxel-based and our planecell methods on KITTI stereo dataset with one frame.}
	\label{Result2}
\end{figure*}

\begin{figure*}[t]
	\centering
	\includegraphics[width=1.0\linewidth]{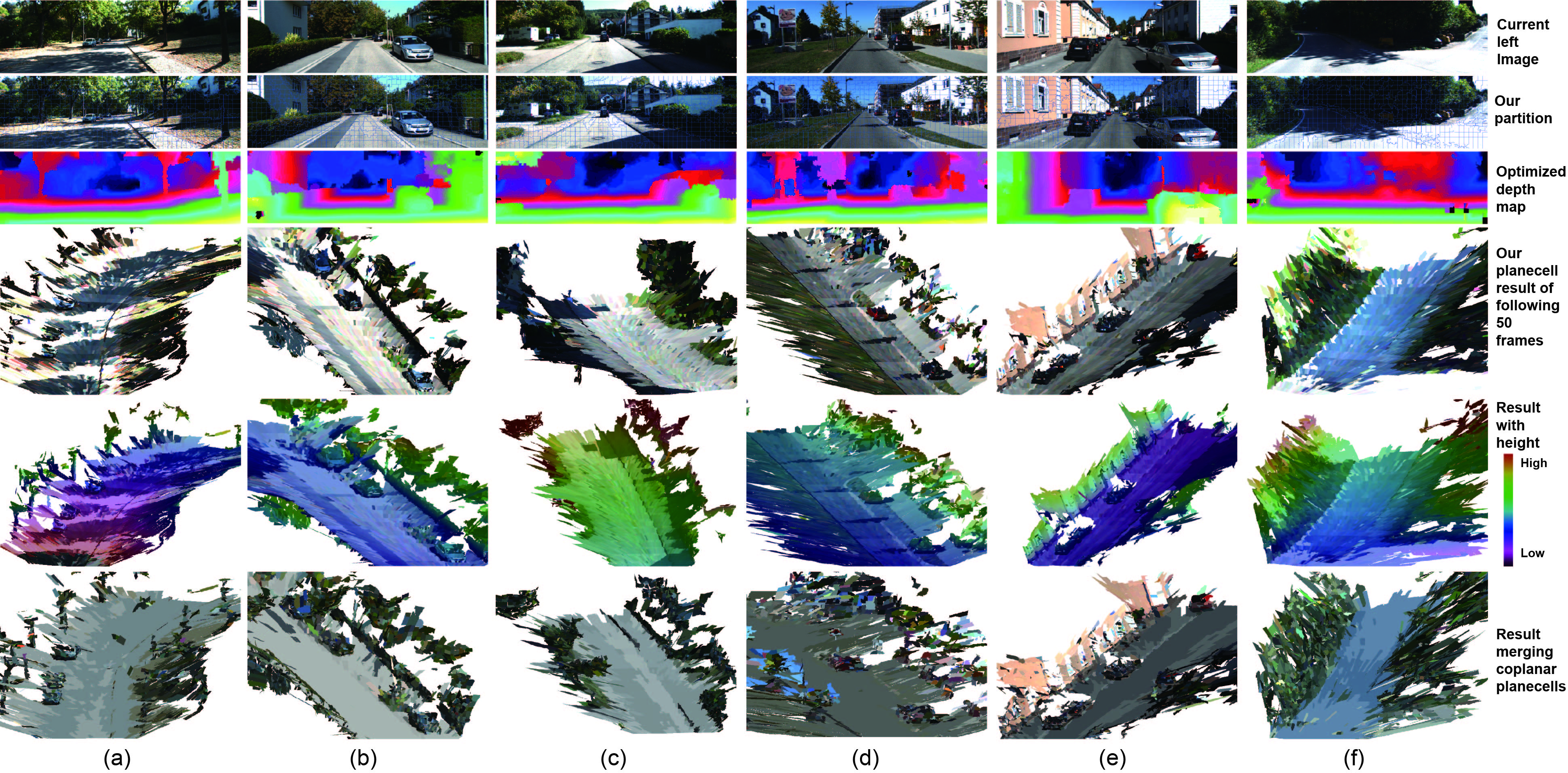}
	\caption{The intermediate steps and reconstruction results of our planecell method from 50 continuous image sequences on KITTI odometry datasets. The results with proposed CRF model aggregating coplanar planecells are also demonstrated.}
	\label{Result1}
\end{figure*}

\section{Experimental Results}
We evaluate our algorithm on three datasets, namely the KITTI stereo dataset, KITTI odometry dataset~\cite{geiger2012we}, and the Middlebury stereo dataset~\cite{scharstein2002taxonomy}. The KITTI stereo dataset separates the images into training and testing sets. The training part includes LiDAR ground truth data with and without occlusions. Each group of images contains two continuous stereo pairs with scene flow information. The outdoor scene dataset provided by the KITTI benchmark is quite challenging, as it contains significant depth variation. Our method shows its advancement dealing with KITTI datasets, whose images are largely of man-made environments that exhibits geometric structures. To better demonstrate the superiority for handling large-scale inputs, we test our algorithm on the KITTI odometry dataset, which has continuous stereo pairs with camera poses. The final dataset on which we evaluate our method is the Middlebury 2001 stereo dataset, which is composed of 9 image groups with ground truths and mostly piecewise planar indoor scenes.

The results are discussed in terms of accuracy, speed, memory requirements, and the ability to represent useful information. We compare our reconstruction accuracy with the point-level method which directly converts 2D pixels into the 3D world. Then, by changing the input depth maps, we test the variation of reconstruction accuracy. We also analyze the 3D map results with a voxel-grid based method. Detailed baselines and evaluations are given in the following.

\subsection{Implementation Details}
As the goal of the proposed method is a new representation of 3D geometric information, we give each planecell an average RGB value for reference. For $\lambda_{reg}$ and $\lambda_{depth}$, we assign them with values according to the initial superpixel size. The plane function is obtained during plane-fitting, and this process may fail if the input depth information is insufficient. In our experiments, the average rate at which the plane function successfully defines all superpixels is $99.95\%$ with the input depth map from SGM. Those planes without plane functions are mostly the area of the sky or reflective objects that will not be converted into the final output. The input depth maps to point or voxel-based 3D space representation methods in our experiments are all calculated by deriving SGM~\cite{hirschmuller2007stereo}. All experiments in this paper only occupy a single core.


\subsection{Baselines}
We compare our results with several state-of-the-art stereo matching algorithms on point-level 3D map reconstruction. The method~\cite{zbontar2016stereo} proposed by Zbontar \etal is a preprocessing step for many stereo algorithms, which utilized a convolutional neural network to calculate the matching cost between patches. The corresponding algorithm named MC-CNN-arct outperforms other approaches on both KITTI and Middlebury stereo datasets. We also compare our results with the matching algorithm of Yamaguchi \etal~\cite{yamaguchi2014efficient} called SPS-st, whose formulation is based on a slanted plane model. Besides, we also test algorithms including SNCC~\cite{einecke2010two}, ELAS~\cite{geiger2010efficient}, SGBM and SGM~\cite{hirschmuller2007stereo} are also listed in our experiments. The SNCC~\cite{einecke2010two} is implemented with additional left-right consistency check and median filters. 

The voxelized representation~\cite{de1999poxels,osman2016patches} is well developed recently for it standardizes the observations of the regions in space. By following this concept, we implement it by dividing the space into 3D voxel grid. The input contains a depth map and a color reference image. The color estimated for each voxel is the average over the observed pixels. The voxel size in our experiments is fixed to $10cm$ for KITTI datasets in our experiments. 

\subsection{Evaluation and Discussion}
We first evaluate reconstruction accuracy by comparing depth maps of each method to ground truth. The sum of per-pixel Euclidean distance errors over the ground truth is computed after reprojecting into the coordinate of left camera. The comparison demonstrates the pixel-level accuracy of our method. Since our method does not lose the position information of each pixel during converting each plane into the 3D space, the comparison is tested on the depth maps. The comparison results are displayed in Fig.~\ref{Compare2Point}. We set the parameters of SPS-st to produce $1000$ superpixels. Our method generates nearly $765$ planecells for KITTI dataset and $270$ plancells for Middlebury dataset referring to the image sizes. For the absence of camera parameters of Middlebury 2001 dataset, we give out the result with the error on the disparities. Ant it can be observed from Fig.~\ref{Compare2Point} that more than $80\%$ points of our results are located within $1m$ around the ground truth. The end of each curve is also restricted to the density of each method.

\begin{table}[t]
	\label{DiffInput}
	\centering
	\begin{small}
		\begin{tabular}{|c|c|c|c|c|}
			\hline
			{Method}&{$<10cm$}&{$<20cm$}&{$<50cm$}&{$<100cm$}\\
			\hline
			\hline
			{SGBM}&{$33.49\%$}&{$50.23\%$}&{$72.91\%$}&{$80.84\%$}\\
			\hline
			{SNCC}&{$57.72\%$}&{$67.06\%$}&{$79.91\%$}&{$85.42\%$}\\
			\hline
			{LiDAR Data}&{$67.56\%$}&{$75.16\%$}&{$85.21\%$}&{$90.73\%$}\\
			\hline
		\end{tabular}
	\end{small}
	\caption{The results on the KITTI stereo dataset by changing the input depth map.}
\end{table}

For the quality of our results depends on the input depth map to some degree, we then test with different inputs in Table.~1. Note that the ground truth from LiDAR can produce planecell model as well. The loss of precision with ground truth inputs is mainly due to inaccurate superpixelization. Another test focuses on changing the number of planecells is shown in Fig.~\ref{Compare2Self}. It demonstrates that the precision increases with more planecells, which is due to the probability of better partition of boundaries. Our \emph{depth term} also helps improve boundary update results. 

We provide several results from three different 3D space representation in Fig.~\ref{Result2}. The input depth maps are all generated by SGM~\cite{hirschmuller2007stereo} method. As shown in Fig.~\ref{Result2}(b) and (e), both point-based and voxel-based results become sparse when the disparities grow, mainly because that far scenes do not have sufficient informations. The proposed planecell method avoids this bad influence by summarizing pixels into a 3D plane which restricts blank area in the output. With the \emph{regularization term} $R$, the partition of our method reduces the complexity of boundaries. The boundaries of each planecell influence both following computation time and storage by defining the vertexes. For the input depth maps generated by SGM~\cite{hirschmuller2007stereo} are not full-dense and include many unmatched areas, the proposed method derives the slanted-plane model to produce optimized depth maps. The planecell also benefits the distance measurements during applications like obstacle avoidance. For instance, denote a position in the 3D world as $(x_0,y_0,z_0)$, the shortest distance to planecell with $\theta '=(A',B',C')$ can be calculated as $|\frac{A'x_0+B'y_0-z_0+C'}{\sqrt{A'^2+B'^2+1}}|$. The proposed method also shows advantages for storing the reconstruction results efficiently. In contrast to point-based method saving all locations, the proposed method requires an average of 45kB per frame.

More detailed results are displayed in Fig.~\ref{Result1} with consecutive frames from KITTI odometry datasets. The reconstruction is based on 50 frames with ground truth poses. The map is reconstructed by mapping each frame data to the first left camera coordinate. Moreover, to show the ability of height perception, we color the placecell with an additional height attribute (see the fifth row of Fig.~\ref{Result1}). The height is an essential variable for path-planning of autonomous driving. With the proposed CRF model, we further aggregate coplanar planecells with plane functions and boundaries. As demonstrated in the last row of Fig.~\ref{Result1}, the coplanar planecells are given the same color. 
In the supplementary materials, we present additional evaluations containing more conditions, like larger-scale reconstructions.

\section{Conclusion}
We propose a novel approach in this paper representing the 3D space with basic units of planes named planecell. The planecells are extracted with a depth-aware manner and can be further aggregated if they belong to the same surface applying proposed CRF model. The experiments demonstrate that our method gives consideration to pixel-level accuracy while efficiently express locations of similar pixels. The results avoid the redundancy of point cloud map and limit output map sizes for further applications. In our future work, we plan to import more complex plane models, like spheres and cylinders, to suit more conditions. We also believe that giving each planecell a semantic label would extend the understanding of the environment in a more effective way.

{\small
\bibliographystyle{ieee}
\bibliography{egbib}
}

\end{document}